\documentclass[10pt, a4paper]{article}
\usepackage{booktabs}
\usepackage{tabularx}
\usepackage[hyphens]{url}

\usepackage[final]{lrec2026} 

\usepackage[normalem]{ulem} 
\usepackage{xcolor}

\title{Training data generation for context-dependent rubric-based short answer grading}

\name{Pavel Šindelář, Filip Prášil, Dávid Slivka, Christopher Bouma, Ondřej Bojar}

\address{Faculty of Mathematics and Physics at Charles University\\
sindelar@ufal.mff.cuni.cz, filip.prasil176@student.cuni.cz, david.slivka745@student.cuni.cz,\\christopher.bouma496@student.cuni.cz, \\
bojar@ufal.mff.cuni.cz
\\}

\abstract{
Every four years, the PISA test is administered by the OECD to test the knowledge of teenage students worldwide and allow for comparisons of educational systems. However, having to avoid language differences and annotator bias makes the grading of student answers challenging. For these reasons, it would be interesting to consider methods of automatic student answer grading. To train some of these methods, which require machine learning, or to compute parameters or select hyperparameters for those that do not, a large amount of domain-specific data is needed. In this work, we explore a small number of methods for creating a large-scale training dataset using only a relatively small confidential dataset as a reference, leveraging a set of very simple derived text formats to preserve confidentiality. Using the proposed methods, we successfully created three surrogate datasets that are, at the very least, superficially more similar to the reference dataset than a straightforward result of prompt-based generation. Early experiments suggest one of these approaches might also lead to improved training of automatic answer grading models.
\\ \newline \Keywords{Short Answer Grading, Optimal Subset Selection, Unlocking Copyrighted Collections} }

\usepackage{cleveref}
\begin{document}

\maketitleabstract

\section{Introduction}
Access to high-quality domain-specific data is commonly one of the biggest obstacles in natural language research, especially for under-explored tasks.

Data for many domains is publicly available in large quantities. However, certain potentially highly scientifically valuable tasks and domains are heavily restricted due to privacy regulations and intellectual property protections. The lack of access to these data restricts the research on it to specific institutions, and also limits the reproducibility of such research.

One example of such highly restricted but interesting domains is the evaluation of short open-ended answers. Training a robust evaluator that can consistently grade open-domain, open-ended answers requires a substantial amount of data from a variety of domains, with diverse question types, and containing both positive and negative examples. It should also include grading rubrics that explain the grading procedures for each question to reduce grading ambiguity. 

Data collection in this domain is often limited by the right to privacy of students. Most openly available datasets fail to meet our criteria. They are typically restricted to a single domain, contain only correct answers, or lack grading rubrics. With the goal of creating the training data for a subtask within the Sensemaking 2026\footnote{\url{https://ufal.mff.cuni.cz/see/sensemaking-2026}} task at the ELOQUENT lab\footnote{\url{https://eloquent-lab.github.io/}}, our research team was granted privileged access to a closed-source dataset containing the relevant information from previous iterations of the Programme for International Student Assessment (PISA) test by OECD. However, because access to this dataset is bound by a non-disclosure agreement, it cannot be published or shared directly.

To address these issues and stay within the bounds of our non-disclosure agreement, we decided to use the confidential data to generate an open proxy dataset to be used instead. However, generating this data is not as straightforward as it might first seem, because using the original data in the generation process directly is difficult and, depending on the method used, could constitute a breach under the NDA. Unfortunately, generating entirely synthetic data would likely produce data that are not very similar to the reference. To mitigate this, we leveraged Derived Text Formats (DTFs). Specifically, we computed a variety of linguistic and semantic features for the reference dataset and the synthetic samples. We then used those features to select the most similar synthetic samples to create the final dataset.
However, since these features are almost entirely focused on the relations between different pieces of text defining a datapoint, it is difficult to compare our method with most other works using DTFs. For example, practically no information about the original datapoints can be recovered from our DTFs without access to parts of the original datapoints.

The code used for subsampling and the full generated dataset can be found in a GitHub repository located at 
\url{https://github.com/sindelarp/training-data-generation-for-context-dependent-rubric-based-short-answer-grading}.

\section{Related works}
There have been several other works on optimal subset selection, often referred to as dataset condensation, distribution matching, or dataset alignment.

GRAD-MATCH
\citep{killamsetty2021grad} selects a subset by approximately minimizing the gradient difference with the validation set for the specific model state at the chosen training step.

LESS \citep{xia2024less} uses a procedure similar to Tracin \citep{pruthi2020estimating} to estimate sample influence on minimizing model loss on a reference dataset by looking at multiple checkpoints during one training run trained on a static dataset. It then selects the samples with the most influence to create a new dataset. 

While potentially less powerful, our method requires significantly less computing power than most of these methods by leveraging the fact that our data has a large number of dependent fields. An added advantage of our method is that the resulting features are potentially shareable. Thus, the generation process can be trivially extended to other datasets we wish to subsample.

The paper \citet{guo2022deepcore} gives an overview of a few methods that are more similar in computational requirements to our method. These, however, deal mostly with monolithic texts and focus on selecting the best subsample without the benefit of a reference dataset, and so models trained on them often have worse results on the test dataset than models trained on the full dataset.

\section{Methodology}

\subsection{Reference Dataset}

The target distribution for our data generation was derived from the dataset of selected questions and answers from the previous iteration of the PISA test, provided to us by the OECD under a non-disclosure agreement. The dataset contained the following fields: 

\begin{itemize}
    \item \textbf{Context:} A text containing all of the relevant context for the question.
    \item \textbf{Question:} A question regarding the context.
    \item \textbf{Grading Rubric:} A grading rubric defining the criteria an answer should meet to be given full, partial, or no credit.
    \item \textbf{Answer:} An answer to the question.
    \item \textbf{Rating:} A ternary rating of the answer, deciding whether it deserves full credit (FC/numerical label 2), partial credit (PC/numerical label 1), or no credit (NC/numerical label 0).
\end{itemize}

Due to the non-disclosure agreement, these items could not effectively be used for anything beyond a closed evaluation. To solve this issue, we decided to treat the dataset as a collection of linguistic and semantic features, which we would then target to mimic in our generated dataset, which would not have such limitations. For the present experiment, we decided to focus only on working with the English subset of the data.

\subsection{Synthetic Candidate Generation}
The initial synthetic dataset had to be large enough to allow for effective sample selection. This was achieved through a multi-stage pipeline that used openly available sources from the Internet and Large Language Model (LLM)\footnote{The open-weights gpt-oss-120b model was used for all parts of the synthetic candidate generation.} prompting.

\subsubsection{Context Extraction}

As the base data for the creation of the contexts, we extracted plain text from several openly accessible websites. These texts were then segmented into sentences using regular expressions. Consecutive sentences were concatenated until they reached a target word count randomly selected between 150 and 800 words. The last sequence for each source was kept only if it contained at least 150 words and was discarded otherwise.

The resulting text chunks were then given to an LLM with the instruction to clean the text of any relics introduced during parsing or transcription. These cleaned texts were then used as contexts.

\subsubsection{Question Generation}

The questions were generated using a 2-stage process. An LLM was given a context and was instructed to generate 1-5 open-ended questions asking about the information from it. It was also instructed to generate questions only satisfying these criteria:

\begin{itemize}
    \item The question is answerable using only the information in the context.
    \item The question is not a subset, non-standalone follow-up question, or rephrasing of any of the previously generated questions.
    \item The question tests the logical understanding of the text, not just factual recall.
    \item The question does not ask for anything subjective. There must always exist a single or only a few correct answers to it.
\end{itemize}

Another LLM was then given all of the generated questions and was tasked with deciding whether each of the given questions follows these criteria. The questions that were identified as compliant were kept; the rest were discarded.

\subsubsection{Grading Rubric Generation}

The grading rubrics were generated in three parts: full credit, partial credit, and no credit. Similarly to question generation, the generation process had two stages. 

In the first stage, an LLM was given a context and a question and was tasked with generating all of the three grading rubric parts so that they follow these criteria:

\begin{itemize}
    \item The full credit section describes a fully correct answer to the question.
    \item The partial credit section describes an answer that is not fully correct (unambiguously), but shows at least partially correct reasoning and understanding of the context and the question.
    \item The no credit section describes an answer that is fully incorrect or shows little to no correct reasoning and understanding of the context and the question.
    \item The individual sections do not have any overlap, which would cause ambiguity when using them to grade.
\end{itemize}

As an additional criterion, we instructed the LLM to make the sections similar in quality and detail to three examples of grading rubrics openly available on the OECD website.

In the second stage, another LLM (the judge) was given the context, question, and the generated grading rubrics and was tasked with deciding whether it follows the generation criteria or not. In the case the text did not follow some criteria, the judge was instructed to generate a critique explaining which criteria it does not follow and why. This explanatory text was then provided along with the instructions from the first stage, and this process was repeated until the judge LLM decided that all of the criteria were satisfied. A limit of 10 iterations was set, but it was never reached.

\subsubsection{Answer Generation and Grading}

The answers were again generated using a 2-stage process. In the first stage, the LLM was instructed to generate five candidate answers for a given context and question, following a given grading rubric section and additional instructions. This was done for each correctness level (full/partial/no credit) separately, with some common and some level-specific instructions. The common instructions were:
\begin{itemize}
    \item The answer follows the given grading criteria.
    \item The answer is not a copy of any of the previously generated answers. It can contain the same information, but it must be worded differently or include/exclude additional information that does not change the correctness of the answer.
    \item Some of the answers may contain common typos and grammatical issues, such as their/they're errors\footnote{We opted to illustrate a spelling error to increase the chances that the model will actually generate some.} or slightly awkward phrasing.
    \item The answers should be written as a high school student might write them under time pressure.
\end{itemize}

During the full-credit-level generation, the model was not given any additional specific instructions.

During the partial-credit-level generation, the specific instruction was:

\begin{itemize}
    \item The incorrectness in the answers comes from a plausible misinterpretation of the text or from simply excluding information required to be fully correct.
\end{itemize}

During the no-credit-level generation, the specific instruction was:

\begin{itemize}
    \item The incorrectness in the answer comes from a plausible misinterpretation of the text, from simply excluding necessary correct information, or trying to "talk around" the question without really answering it.
\end{itemize}

Similarly to the first stage, the second stage was also split by the correctness levels. For each correctness level, an LLM judge was given the context, the question, the grading rubric section, the instructions that the generation process of the answers was supposed to follow, and the answers themselves. Its task was to decide whether each of the answers was generated by correctly following the generation instructions. The answers for which this was not the case were discarded.

\subsection{Feature Extraction}

To better understand the generated dataset, we extracted the following features for each datapoint:

\begin{itemize}
    \item \textbf{BAAI/bge-m3\footnote{\url{https://huggingface.co/BAAI/bge-m3}} context question cosine similarity}
    is the cosine similarity between dense embeddings of the context and the question, where embeddings are produced by the \textbf{BAAI/bge-m3} encoder.
    
    \item  \textbf{BAAI/bge-m3 context answer cosine similarity}
    is the cosine similarity between \textbf{BAAI/bge-m3} embeddings of the context and the answer.

    \item \textbf{BAAI/bge-m3 rubrics/FC answer cosine similarity}
    is the cosine similarity between \textbf{BAAI/bge-m3} embeddings of the full-credit rubric text and the student answer.

    \item \textbf{recall 2gram FC answer}
    is the number of shared bigrams between the full-credit rubric and the answer divided by the length of the answer.    The value \textbf{recall 2gram NC answer}
    is similar, but uses the no-credit rubric.

    \item \textbf{answer length}
    is the length of the answer measured as the number of tokens.

    \item \textbf{answer lexical density}
    is the proportion of answer tokens tagged as content words (nouns, verbs, adjectives, adverbs) by NLTK POS tagging.
    
    \item \textbf{Jaccard 1gram question answer} and \textbf{Jaccard 1gram context answer}
    are the Jaccard similarity between the unigram sets of the question and the answer and between the unigram sets of the context and the answer, respectively.

    \item \textbf{recall 2gram question answer} and \textbf{recall 2gram context answer}
    are the fraction of answer bigrams that also appear in the question and answer bigrams that also appear in the context, respectively.
    
    \item \textbf{recall 2gram context overlap with answer minus question}
    is a recall measure. Let $S$ be the set of n-grams in the answer with n-grams from the question removed (set difference). This feature is the fraction of n-grams in $S$ that also occur in the context.
        
    \item \textbf{tfidf cosine question answer} and \textbf{tfidf cosine context answer}
    are the cosine similarity between TF--IDF vector representations of the question and the answer and of the context and the answer, respectively.

    \item \textbf{precision 1gram question answer}
    is the fraction of question unigrams that also appear in the answer.
    
    \item \textbf{recall 1gram question answer} and \textbf{recall 1gram context answer}
    are the fraction of answer unigrams that also appear in the question and answer unigrams that also appear in the context, respectively.
    
    \item \textbf{recall 1gram context overlap with answer minus question}
    is a recall measure similar to \textbf{recall 2gram context overlap with answer minus question}.
\end{itemize}

\subsection{Data selection}
To make the data resemble the OECD data more closely, we computed the features mentioned above for each datapoint in the synthetic candidate dataset and the OECD dataset. We then compared three feature-based matching methods.

\paragraph{First selection method} In the \textbf{first selection method}, we use a very small dataset and simply take the 5\% of samples that are the most similar to the mean of the OECD features of their label (i.e., we only compare FC with FC, PC with PC, and NC with NC) under the L2 metric.

\paragraph{Second selection method} As shown in the section below the \textbf{first selection method} did not yield any noticeable improvement in \textbf{E1} and \textbf{E2}. We identified two potential issues we tried to alleviate in the \textbf{second selection method}.
\begin{enumerate}
    \item The lack of domain coverage after subsampling. -- We take the best 5\% in every domain separately to try to maintain data diversity. For the few-shot examples, we sample 2 datapoints from the subsampled data for each domain.
    \item The potential of the single mean being too reductive. -- Instead of looking at the similarity to a single mean over all OECD data, we decided to look at the similarity to the closest of eight different representatives of the OECD data feature space. These representatives were selected using k-means with k-means++ initialization.
\end{enumerate}

\paragraph{Third selection method} In the \textbf{third selection method}, we tried to further relax the conditions. Instead of demanding that the features not reveal information about individual samples, we used access to the unreduced features to select the training dataset. We decided to remove the direct reliance on distance comparisons and instead look at the rank of the first five closest datapoints from the reference dataset. To get the negative score of the sample $a$, we sort the reference dataset and the generated dataset together by their L2 distance to $a$, ascending, and average the indices of the first five samples belonging to the reference dataset in this ordering.

\section{Results}
We decided to compare the usefulness of the subsampled dataset to the full dataset using multiple comparison methods.

\paragraph{\textbf{E1}} For a fast preliminary experiment, we decided to try to see how different the respective dataset utility is by sampling few-shot examples from both the full dataset and the dataset sub-sampled via the selection method and comparing the accuracy of \textit{Qwen3-8B} using these two different parameters on the confidential data. We used a simple prompt generated using DSPy \citep{khattab2022demonstrate}.

\paragraph{\textbf{E2}} We also decided to do a simple test on how good our subsampled subset would be for model selection. If the results of metrics on the selected subset are closer than on the whole dataset, it could be expected that if the subset was used to select between different model architectures, different models, training hyperparameters, or different model checkpoints, it would lead to better results on the confidential dataset than selecting based on the full dataset.

\paragraph{\textbf{E3}} For additional comparison, we chose to train a small model on the dataset resulting from the application of the selection method. To make the small size of the dataset work with the model and make the training more stable, we chose to use the "tasksource/ModernBERT-base-nli" as a base for our training. This model is based on the pretrained model ModernBERT and further finetuned on a mix of natural language inference datasets. As a sort of regularization, we added a subset of the 20\% samples with the longest tokenized length from the MNLI dataset. We used a learning rate of 2e-5 and did gradient accumulation over 32 samples in each step. These were selected by training on an unrelated training dataset and scoring on a manually verified development dataset, where the model achieved a mean accuracy of 0.722 during the second half of training. We decided to treat the evaluation on the confidential dataset during training as a set of paired experiments. We evaluated the model trained on a randomly subsampled dataset and a model trained on a dataset subsampled using the method after every 187 steps, and then calculated the Wilcoxon signed-rank test statistic \citep{wilcoxon1992individual}. We note that while consecutive accuracies of the model on the reference set during training are not independent, they are approximately independent enough for an initial estimate. Ideally, we would do many independent runs and analyze accuracy at each number of steps separately.

\subsection{First selection method}
The feature similarity results of \textbf{first selection method} can be seen in \cref{stats_improvement}, the improvement was especially significant when it came to answer length, which is, of course, a trivial feature. 

On the other hand, some features, such as the bigram recall between the no credit rubric and the answer, now differ more. For this particular example, this is most likely caused by the fact that the no-credit conditions were often described abstractly.

\begin{table}[!ht]
\centering
\footnotesize
\begin{tabularx}{\columnwidth}{X|r|r}
\toprule
& original & selected \\
\midrule
BAAI/bge-m3 context question cosine similarity & 0.009 & 0.026 \\
BAAI/bge-m3 context answer cosine similarity & 0.047 & 0.009 \\
BAAI/bge-m3 rubrics/FC answer cosine similarity & 0.074 & 0.053 \\
recall 2gram FC answer & 0.045 & 0.087 \\
recall 2gram NC answer & 0.005 & 0.028 \\
answer length & 30.691 & 16.987 \\
answer lexical density & 0.030 & 0.014 \\
jaccard 1gram question answer & 0.054 & 0.024 \\
jaccard 1gram context answer & 0.015 & 0.001 \\
recall 2gram question answer & 0.009 & 0.017 \\
recall 2gram context answer & 0.143 & 0.130 \\
recall 2gram context overlap with answer minus question & 0.157 & 0.140 \\
tfidf cosine question answer & 0.101 & 0.046 \\
tfidf cosine context answer & 0.025 & 0.014 \\
precision 1gram question answer & 0.197 & 0.110 \\
recall 1gram question answer & 0.028 & 0.055 \\
recall 1gram context answer & 0.141 & 0.093 \\
recall 1gram context overlap with answer minus question & 0.142 & 0.083 \\
\bottomrule
\end{tabularx}
\caption{Absolute differences between the overall feature means between confidential data and datasets.}
\label{stats_improvement}
\end{table}

As we can see in \cref{tab:comparison_of_scores}, our method did not improve model performance in comparison method \textbf{E1}. This could be because \textbf{first selection method} is too coarse for the selected model to benefit from the selected few-shot data more.

We can see in \cref{tab:comparison_of_scores} that when we used method \textbf{E2}, the results on the selected data seem to be overall better than on both the entire dataset and on the confidential data, indicating it would likely not be a good dataset for model selection.

This indicates that \textbf{first selection method} is not strong enough to balance out the specific biases this dataset elicits in our model.

\begin{table}[!ht]
\centering
\footnotesize
\begin{tabularx}{\columnwidth}{lXrr}
\toprule
fewshot & eval  & Accuracy  & Quadratic \\
source & target & &  Weighted\\
 &  & & Kappa\\
\midrule
All & Confidential & 0.000 & 0.000 \\
Selected & Confidential & -0.040 & -0.010 \\
All & Selected & +0.250 & +0.380 \\
All & All & +0.120 & +0.410 \\
\bottomrule
\end{tabularx}
\caption{Comparison of performance measures. Note that the accuracy was calculated on a dataset resampled to be balanced. Quadratic Weighted Kappa is computed using the respective numerical labels.}
\label{tab:comparison_of_scores}
\end{table}

\subsection{Second selection method}
We later tried to evaluate the \textbf{second selection method} on a larger dataset. The results are very similar to the results of \textbf{first selection method}, and can be seen in \cref{tab:comparison_of_scores2}.

Again, there seems to be a small decrease in accuracy, 0.58 vs 0.61, in \textbf{E1}. This could be because our method does not maintain diversity. When comparing how useful the data from \textbf{second selection method} is for model selection. We can see in the results of \textbf{E2} in  \cref{tab:comparison_of_scores2} that the accuracy on the selected subset, 94 vs 92, is very similar to the accuracy on the whole dataset.

\begin{table}[!ht]
\centering
\footnotesize
\begin{tabularx}{\columnwidth}{lXrr}
\toprule
fewshot & evaluation & Accuracy & Quadratic \\
source & target & & Weighted\\
 &  & & Kappa\\
\midrule
All & Confidential & 0.000 & 0.000 \\
Selected & Confidential & -0.033 & -0.019 \\
All & Selected & +0.333 & +0.469 \\
All & All & +0.313 & +0.456 \\
\bottomrule
\end{tabularx}
\caption{Comparison of performance measures on the second method, see \cref{tab:comparison_of_scores}.}
\label{tab:comparison_of_scores2}
\end{table}

As an initial estimate using \textbf{E3}, we got the high $p$ of $0.715$, significantly above the threshold of $0.05$. This suggests further experiments would likely be fruitless.

\subsection{Third selection method}
\begin{figure}
    \centering
    \includegraphics[width=0.75\linewidth]{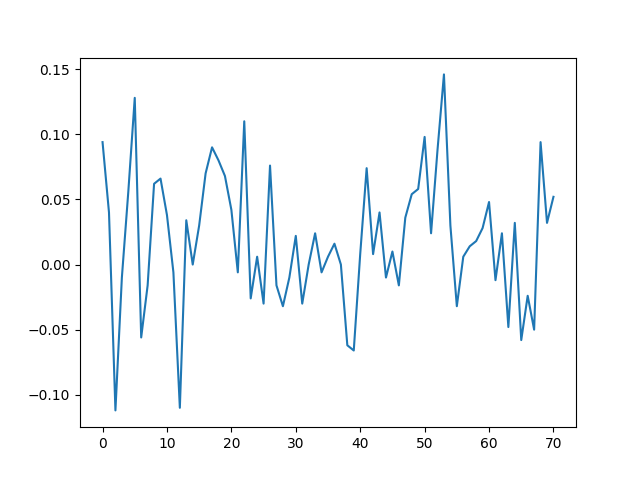}
    \caption{Accuracy of the model training in the subsampled dataset minus the accuracy of the model training on a randomly subsampled dataset.}
    \label{accuracy_difs}
\end{figure}

Because we found it the most rigorous, we only evaluated the third selection using \textbf{E3}, the differences used for the test can be seen in \cref{accuracy_difs}. The result was a $p$ of $0.00091$, below the threshold of $0.05$. Over the training, the model on the subselected data achieved an average gain of $0.0206$ accuracy.

Afterwards, we reran the experiments twice more for each dataset with a different order of the training set. Since the weights of all layers were kept from the base model, this was the only way of randomizing the training run we attempted.

Here, we did a Wilcoxon signed-rank test by selecting a pair of training runs to compare separately at each step of evaluation, using each accuracy value exactly once. The result was a $p$ of $0.00102$, below the threshold of $0.05$. Over the training, the models on the subselected data achieved an average gain of $0.0116$ accuracy.  

\section{Conclusion}
We presented a method for deriving a dataset of contexts, questions, scoring rubrics, and scored answers from a closed seed set, namely the OECD PISA test items. We generate these items using an open-weight LLM and compare different methods of optimal subset selection.

Based on our experiments, only the most relaxed method (\textbf{third selection method}) has achieved significant results. This direction of research seems promising, but determining whether such methods truly select useful data for the purposes of training in general would require more thorough experiments on many different models.

In the future, we would like to iterate on the subset selection methods that use reduced features. It seems sets of simple, averaged features are often not expressive enough, or they need to be weighted using a complex set of hyperparameters, which would need to undergo a resource-intensive selection process. Likely, a more expressive set of features will be necessary. These methods might work better on more heavily structured datasets.

\section{Limitations}
Our evaluations generally work with only a single run, and a more rigorous repetition of the experiments on many different datasets and models would be necessary to test the potential of the method.

\section{Declaration on Generative AI}
LLMs were used for the purposes of a preliminary search for style and grammar errors. 
This work contains no LLM-generated text.

\section{Acknowledgements}

The authors acknowledge the funding from the Project OP JAK Mezisektorová spolupráce Nr. CZ.02.01.01/00/23\_020/0008518 named ``Jazykověda, umělá inteligence a jazykové a řečové technologie: od výzkumu k aplikacím'', the support of the National Recovery Plan funded project MPO 60273/24/21300/21000 CEDMO 2.0 NPO and the support by EC Digital Europe Programme (DIGITAL) grant number 101195233 (OpenEuroLLM).

Computational resources were provided by the e-INFRA CZ project (ID:90254), supported by the Ministry of Education, Youth and Sports of the Czech Republic.

\section{Bibliographical References}\label{sec:reference}

\bibliographystyle{lrec2026-natbib}
\bibliography{lrec2026-example}

\bibliographystylelanguageresource{lrec2026-natbib}
\bibliographylanguageresource{languageresource}

\end{document}